\documentclass{article}

\usepackage{PRIMEarxiv}
\usepackage{cite}
\usepackage{amsmath,amssymb,amsfonts}
\usepackage{algorithmic}
\usepackage{textcomp}
\usepackage{xcolor}
\usepackage{subcaption}
\usepackage{balance}
\usepackage{todonotes}
\usepackage{multirow}%
\usepackage{booktabs}
\usepackage{comment}
\usepackage[utf8]{inputenc} 
\usepackage[T1]{fontenc}    
\usepackage{hyperref}       
\usepackage{url}            
\usepackage{booktabs}       
\usepackage{amsfonts}       
\usepackage{nicefrac}       
\usepackage{microtype}      
\usepackage{lipsum}
\usepackage{fancyhdr}       
\usepackage{graphicx}       
\graphicspath{{media/}}     
\def\BibTeX{{\rm B\kern-.05em{\sc i\kern-.025em b}\kern-.08em
    T\kern-.1667em\lower.7ex\hbox{E}\kern-.125emX}}
\title{Enhancing Atrial Fibrillation Detection Through Hyperparameter Optimization}

\author{
  Sara Gjorgjieva$^{1}$ $^{2}$ \href{https://orcid.org/0009-0009-2014-4515}
    {\raisebox{0.8ex}{\scriptsize[0009-0009-2014-4515]}},  
  Ilinka Ivanoska$^{2}$
    \href{https://orcid.org/0000-0002-9363-1527}
    {\raisebox{0.8ex}{\scriptsize[0000-0002-9363-1527]}},\\
  Marjan Gusev$^{1}$ $^{2}$ 
    \href{https://orcid.org/0000-0003-0351-9783}
    {\raisebox{0.8ex}{\scriptsize[0000-0003-0351-9783]}}\\
  \\
  $^{1}$Innovation Dooel, Skopje, North Macedonia \\
  $^{2}$Ss. Cyril and Methodius University in Skopje, Faculty of Computer Science and Engineering, North Macedonia \\
  \\
  \texttt{sara.gjorgjieva@innovation.com.mk} \\
  \texttt{\{ilinka.ivanoska, marjan.gushev\}@finki.ukim.mk}}

\begin{document}
\maketitle

\vspace{-2em}
\begin{center}
\small
In \textit{ICT Innovations 2025. AI and the Digital Frontier},\\ Communications in Computer and Information Science series, Springer (In Press).
\end{center}
\vspace{2em}

\begin{abstract}
This study explores the role of hyperparameter optimization (HPO) in enhancing the detection of Atrial Fibrillation (AF) using both machine learning and deep learning models. We evaluated a range of standard classifiers - Decision Tree, Random Forest, and XGBoost classifiers, alongside deep learning architectures including Recurrent Neural Networks, Deep Neural Networks, and Siamese Neural Networks. To fine-tune these models, we applied four HPO methods: Genetic Algorithm, Hyperband, Optuna, and SMAC3. The experiments were conducted using inter-patient and intra-patient data splits to assess model generalization and robustness in the presence of class imbalance, which is reflective of real-world AF occurrence. Our results, based on PhysioNet electrocardiogram datasets, showed that the Hyperband-optimized XGBoost model outperformed all others, achieving a peak F1 score of 96.95\%. These findings highlight the critical role of suitable HPO strategies in enhancing the accuracy and robustness of AF classification outcomes in clinical settings.
\end{abstract}


\section{Introduction}
\label{introduction}

Machine learning (ML) has revolutionized various domains by offering innovative solutions to real-world challenges, improving the quality of life, and extending life expectancy. Its versatility stems from the fact that ML models are highly adaptable and can be applied across diverse fields, regardless of the subject matter. However, selecting the most suitable model for a problem can be challenging, requiring substantial time, resources, and expertise.

Once a model is selected, the next critical step is optimizing its performance. This involves fine-tuning the model parameters to ensure it delivers the best results for the given task. Model effectiveness is typically evaluated based on its ability to generalize and perform with high accuracy to unseen data, striking a balance between underfitting and overfitting. Achieving this balance depends on two key factors: the model features and the hyperparameters that dictate the model's learning behavior \cite{weerts2020importance,aliferis2024overfitting}.
Although related, hyperparameters and parameters differ significantly in their roles. Parameters, such as weights and coefficients, are determined by the model during training and represent how predictor variables influence the target variable. On the other hand, hyperparameters are external configuration variables set before training and control the model's structure, learning rate, and other operational aspects. While default hyperparameter values exist, they often need to be tailored to the specific dataset and model through experimentation, known as hyperparameter tuning or hyperparameter optimization (HPO).
HPO involves systematically testing different hyperparameter combinations to identify the one that achieves the best performance. Unlike parameters adjusted during training, hyperparameters are predefined and remain fixed throughout training. Choosing the correct hyperparameter values directly impacts model performance, making hyperparameter tuning a critical step in the ML pipeline \cite{feurer2019hyperparameter, nyuytiymbiy2020parameters}.
Data scientists usually enhance a model's predictive capabilities by fine-tuning hyperparameters and evaluating the performance of new data. This study focuses on the importance of HPO in improving the detection of atrial fibrillation (AF), and careful experimentation is planned to achieve the best results. We evaluate the effect of various HPO techniques: Hyperband \cite{li2018hyperband, bischl2023hyperparameter} , Optuna \cite{akiba2019optuna, agrawal2020optuna}, Genetic Algorithm (GA) \cite{lambora2019genetic, katoch2021review}, and Sequential Model-Based Algorithm Configuration (SMAC3)  \cite{lindauer2022smac3, rook2025mo, truong2019towards}, highlighting their unique advantages.

The main research question is to evaluate selected HPO methods for AF detection models, focusing on robustness and the generalizability of optimized hyperparameters across different patient distributions and benchmark datasets. Unlike previous studies, our work explicitly incorporates inter- and intra-patient evaluations to assess robustness more thoroughly, where intra-patient refers to training and testing within the same dataset, and inter-patient refers to training on one dataset and testing on another.

Section~\ref{introduction} outlines the significance of AF detection and the role of HPO in enhancing model performance. Section~\ref{related-work} reviews prior work on AF detection using ML. Section~\ref{methodology} describes the datasets, models, and HPO techniques applied in this study. Section~\ref{setup} presents the experimental setup, implementation tools, execution time, and model performance results. Section~\ref{discussion} compares our findings with prior research and evaluates the strengths and limitations of the HPO methods used. Finally, Section~\ref{conclusion} summarizes the key insights and suggests directions for future work.


\section{Related Work}
\label{related-work}
Several studies have applied HPO techniques to improve AF detection with various ML models. A common limitation is the lack of explicit handling of class imbalance—most optimize for accuracy, which biases models toward the majority class. Without F1-score or AUC, tuning often yields poor recall for AF cases, reducing robustness. Our study addresses this by emphasizing metrics that better reflect real-world detection challenges.

Optuna has been widely adopted for AF detection due to its flexibility. Tudjarski et al. \cite{tudjarski2021new} used Optuna with XGBoost on 41-beat windows, achieving 99\% F1-score. Rawshani et al. \cite{rawshani2024enhancing} applied it in a multimodal framework combining CNNs (AlexNet, VGG-16) with XGBoost, reporting 97.23\% AUC-ROC. However, both relied on simple splits (80/20, 70/30) with mixed patient samples, limiting generalizability. By contrast, our work uses multiple datasets and patient-wise splits for broader robustness.

GA has also shown promise in complex search spaces. Al Qaraghuli et al. \cite{al2022detection} fine-tuned a CNN using GA, achieving 100\% sensitivity, 98.90\% specificity, and 99.95\% accuracy on MIT-BIH AF. Yet, unclear data-splitting raises concerns about overfitting.

Hyperband has been applied to DL optimization by balancing exploration and exploitation. Han et al. \cite{han2020optimal} used it on CNN and LSTM, achieving 72.80\% accuracy and 59.31\% macro-F1, slightly outperforming manual tuning. Chen et al. \cite{chen2024identifying} optimized a Dual-Input Mixed Neural Network (DMNN) with ECG and demographic data, reaching 83.4\% AUC. While Han et al. used 5-fold CV on CinC 2017, Chen et al. used two datasets but only one experiment from a single institution, risking demographic bias.

Our review found no prior use of SMAC3 for AF detection.

While all HPO methods improve performance, their strategies and efficiency differ. Hyperband is efficient for DL with limited resources; Optuna explores structured spaces via probabilistic modeling; GA handles high-dimensional searches but is computationally expensive. The choice depends on model complexity, dataset properties, and resources. Importantly, these methods do not inherently address imbalance—bias often arises from accuracy-based tuning. To counter this, we use F1-score and AUC as optimization metrics for more reliable performance on AF datasets.

\section{Methodology}
\label{methodology}

\subsection{Datasets}
\label{datasets}

We used two publicly available ECG datasets from PhysioNet \cite{moody2022physionet}: the MIT-BIH Arrhythmia Database (MITDB) \cite{moody1992bih} and the Long-Term AF Database (LTAFDB) \cite{biczuk2024prr30}. MITDB contains 30-minute ECG recordings from 48 patients, with over 100,000 annotated heartbeats, approximately 10\% correspond to AF. LTAFDB offers 24 hours recordings from 83 patients, with a higher AF prevalence of about 50\%. Due to its large size, we used a subset of around 940,000 annotated heartbeats, obtained by segmenting the recordings into fixed-length windows of 42 beats across all patients.

\begin{figure}[!h]
\centering
\includegraphics[width=0.25\textwidth]{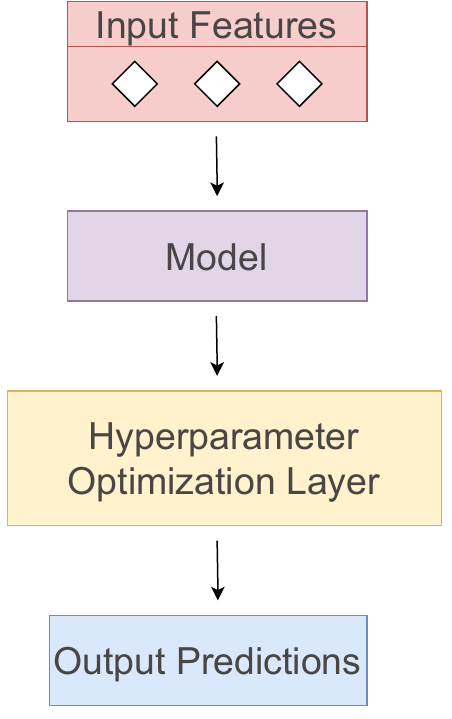}
\caption{Overview of the HPO-supported ML model development}
\label{fig2:model_architecture}
\end{figure}

\subsection{Models}

To effectively detect AFIB, we explored a range of models spanning from traditional machine learning to more advanced neural architectures.
\textbf{Decision Trees (DT)} \cite{bin2017detection} serve as a straightforward baseline, offering quick and interpretable classifications by recursively splitting features, though their simplicity often limits their ability to model complex ECG dynamics.\\
Building on this, \textbf{Random Forests (RF)} \cite{zabihi2017detection} introduce ensemble learning by combining multiple DTs trained on random subsets of the data, which improves robustness and reduces the risk of overfitting—particularly useful for handling noise in ECG recordings. \textbf{XGBoost (XGB)} \cite{chen2018classification} further enhances tree-based modeling by applying gradient boosting with built-in regularization, allowing it to capture subtle patterns in structured heartbeat features while maintaining generalization. \\
In contrast to these tree-based methods, \textbf{Recurrent Neural Networks (RNNs)} \cite{shashikumar2018detection} are well-suited for sequential data like ECG signals, as they can model temporal dependencies between heartbeats. Our implementation of RNNs includes techniques like dropout, L2 regularization, and Leaky ReLU activations to improve learning stability and prevent overfitting. \textbf{Deep Neural Networks (DNNs)} \cite{raghunath2021deep}, while not inherently sequential, extract hierarchical features through stacked dense layers and are particularly effective when working with engineered inputs; regularization and dropout here also support better generalization.\\ Finally, \textbf{Siamese Neural Networks (SNNs)} \cite{ng2023few} take a unique approach by learning to compare pairs of ECG segments using twin networks with shared weights. This architecture excels in tasks that require identifying subtle rhythm similarities or differences—making it especially powerful for distinguishing AFIB from normal sinus rhythm in a patient-agnostic way.

\subsection{Hyperparameter Optimization Techniques}
\label{hpo-techniques}

In our previous work \cite{gjorgjieva2024af}, we manually set the hyperparameters for these models. However, manual tuning is both time-consuming and inefficient, often failing to identify optimal parameter combinations for best performance.

Even though there is no single HPO method or universal set of hyperparameters that guarantees optimal performance across all models and datasets. Therefore, a systematic and tailored approach is essential. The key challenge lies in selecting the next combination of hyperparameters to test the exploration of new configurations, in order to consistently achieve better performance.

To ensure diversity in optimization strategies, we selected four HPO techniques that employ fundamentally different search mechanisms, allowing us to compare their effectiveness in improving AF detection: Hyperband \cite{chen2024identifying} , Optuna  \cite{tudjarski2021new} , \cite{rawshani2024enhancing}, GA \cite{al2022detection}, and SMAC3. Each of these approaches offers different optimization strategies and is suited to specific scenarios, providing a robust framework for improving model performance in detecting AF.

Fig.~\ref{fig2:model_architecture} provides an overview of the architecture used to optimize ML model development through HPO techniques.
\\
In the following sections, we briefly describe how each technique functions and why it is relevant for optimizing ML models in the context of AF classification.

\textbf{Hyperband} is a powerful HPO technique that uses bandit-based early stopping to efficiently allocate computational resources. It is especially beneficial when working with complex time-series models, such as those used in AF detection, as it quickly eliminates poorly performing hyperparameter configurations and prevents excessive resource consumption. In this domain, where models process large volumes of ECG signal data, Hyperband ensures that only the most promising configurations are further explored. This dynamic resource allocation accelerates the tuning process for computationally intensive models like RNNs and DNNs, helping improve both training efficiency and performance.

\textbf{Optuna} is a modern HPO method that uses Bayesian Optimization and Tree-structured Parzen Estimators (TPE) to efficiently search for optimal hyperparameters. It intelligently balances exploration and exploitation, making it more effective than grid or random search in high-dimensional spaces. In AF detection, where fine-tuning impacts critical aspects like feature extraction and model sensitivity, Optuna’s built-in pruning mechanism is especially useful, it can stop unpromising trials early, reducing training time without compromising performance. This is particularly advantageous when optimizing complex models like neural networks or XGBoost.

\textbf{Genetic Algorithm} (GA) mimics the process of natural selection to evolve better hyperparameter combinations across generations. GA’s population-based search is beneficial for AF detection models due to its ability to explore a wide and diverse range of configurations. Techniques like mutation and crossover help GA escape local optima, making it well-suited for tuning highly nonlinear models like DNNs and SNNs. This is particularly important when working with datasets such as MITDB and LTAFDB, where data distributions vary and require adaptive tuning for optimal classification.

\textbf{SMAC3}(Sequential Model-Based Algorithm Configuration) is a Bayesian optimization approach that builds a probabilistic model of the search space and iteratively refines it to guide the search. For AF detection tasks, where models must capture subtle ECG signal variations, SMAC3 offers a systematic and data-driven method to identify high-performing hyperparameter configurations. It is especially useful for tree-based models like RF, DT, and XGB, as it can efficiently tune hyperparameters such as depth, learning rate, and feature criteria. By learning from past evaluations, SMAC3 reduces redundant trials and focuses computational effort on the most promising regions of the search space.

%

%
\section{Experimental Setup and Results}
\label{setup}
\subsection{Experiments}
\label{experiments}

To evaluate the effectiveness of HPO, we designed four experimental setups that vary in training and testing data splits. These experiments are categorized into \textbf{intra-dataset} (same dataset for training and testing) and \textbf{inter-dataset} (different datasets for training and testing) evaluations. Only clean AF segments were used for training, while testing included all signal types.\\

\textbf{Intra-dataset evaluations} use an 80/20 split:
\begin{itemize}
    \item []\textbf{E11}: MITDB is used for both training and testing.
    \item []\textbf{E22}: LTAFDB is used for both training and testing.
\end{itemize}

\textbf{Inter-dataset evaluations} involve cross-dataset testing:
\begin{itemize}
    \item []\textbf{E12}: Train on MITDB, test on LTAFDB.
    \item []\textbf{E21}: Train on LTAFDB, test on MITDB.
\end{itemize}

These setups allow us to assess model robustness across intra-patient and inter-patient scenarios. In E11 and E22, samples from the same patients are kept separate between training and testing (intra-patient split). In E12 and E21, datasets differ completely (inter-patient setting), helping evaluate generalization across patient populations.

Table~\ref{tab-combinations} summarizes the experimental configurations applied to each HPO method.

\begin{table}[!h]
\centering
\caption{Specification of experiments}
\label{tab-combinations}
\begin{tabular}{@{}lllll@{}}
\toprule
 \textbf{Exp.} & \textbf{Train} & \textbf{Test} & {\bf Evaluation}	&  {\bf Split } \\ 
\midrule
E11	& MITDB	& MITDB	& intra-patient	& 80/20  \\
E12	& MITDB	& LTAFDB	& inter-patient	& dataset\\
E21	& LTAFDB	& MITDB	& inter-patient	& dataset  \\ 
E22	& LTAFDB	&LTAFDB	& intra-patient	& 80/20  \\ 
\hline
\end{tabular}
\end{table}

\subsection{Implementation}
\label{implementation}

All experiments were implemented in Python using Scikit-learn, XGBoost, TensorFlow, and PyTorch. For hyperparameter optimization, we applied Optuna, GA (DEAP), Hyperband (HpBandSter), and SMAC3.

Training was conducted on an AMD Ryzen 7 4800H CPU, leveraging all cores for parallelization. No GPU was used, and computations relied on the default system configuration. Each method ran 50 trials, with the F1-score as the optimization objective to balance precision and recall.

Table~\ref{tab-configuration_metrics} summarizes the configurations of the HPO techniques. Parameter settings were chosen based on prior work, recommendations from the respective libraries, and empirical testing on smaller pilot runs. For example, Hyperband parameters were tuned to allow early stopping without sacrificing exploration; Optuna was capped at 50 trials to ensure convergence within the available budget; GA settings balanced population size and mutation rate to explore both shallow and deeper evolutionary searches; and SMAC3 emphasized quality-driven optimization to avoid overfitting to rapid gains. These choices reflect trade-offs between computational efficiency, robustness, and thoroughness in exploring the hyperparameter space.

\begin{table}[!h]
\centering
\caption{HPO Models Configurations}
\label{tab-configuration_metrics}
\begin{tabular}{@{}lll@{}}
\toprule
 \textbf{Optimizer} & \textbf{Parameter} & \textbf{Value}\\
\midrule
        &  factor   & 3\\ 
        Hyperband & max resources   & 100 \\ 
 	& min resources  & 20 \\
  	& max epoch  & 20 \\        \midrule 
  	&   &  \\ 
        Optuna & trials  & 50 \\
 	&   &  \\        \midrule
  	& population & [10,30,70]\\ 
        GA & ngen  & [5,10,30] \\ 
 	& cxpb & 0.5 \\
 	& mutpb  & 0.2 \\        \midrule
 	&   &  \\ 
         SMAC3 & trials   & [30,50,150]\\ 
 	& objectives  & quality \\         
\hline
\end{tabular}
\end{table}

%
\subsection{Execution Time Results}
\label{sec:time-execution}

\begin{figure}[!ht]
\centering
\includegraphics[width=\textwidth]{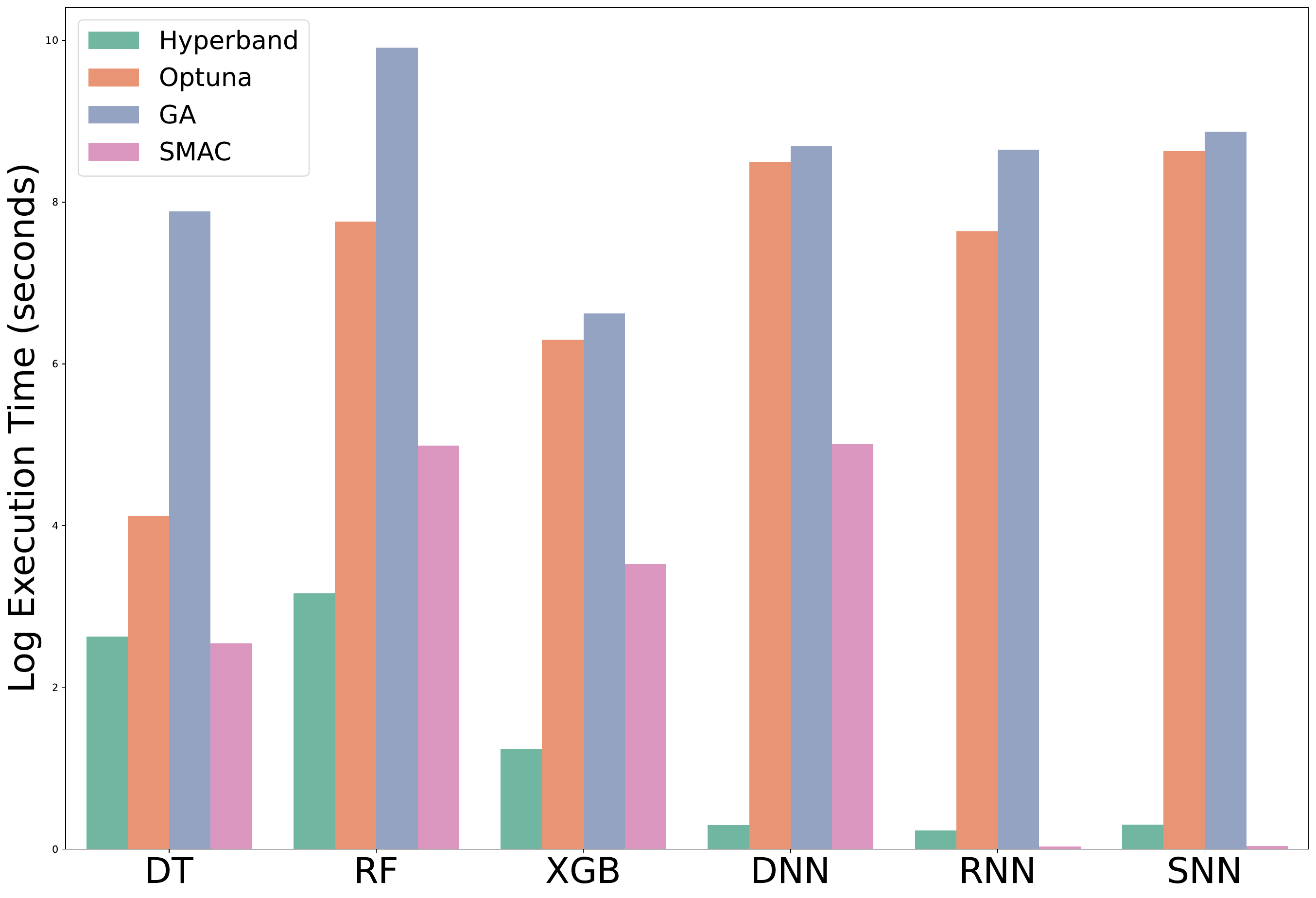}
\caption{Visual representation for the execution time(s) for different HPO}
\label{fig:execution_time}
\end{figure}

Fig.~\ref{fig:execution_time} presents the execution times (in seconds on a logarithmic scale) for all experiments applying different HPO techniques. Hyperband exhibits the lowest execution time for the neural networks, while SMAC3 performs the fastest for the RNN and SNN. Overall, Hyperband and SMAC3 are the most computationally efficient HPO methods.

In contrast, GA has the highest execution times, particularly for complex models like RF, DNN, RNN, and SNN, indicating its high computational cost. Optuna is also relatively expensive for deep learning architectures. These results reflect differences in optimization strategies, where Hyperband and SMAC3 achieve efficiency through early stopping and surrogate modeling, while Optuna and GA rely on more exhaustive search procedures.

%

\subsection{Performance Results}
\label{experimental-findings}

Given the class imbalance typical in AF detection, relying solely on accuracy can be misleading. Therefore, we report multiple metrics—Precision, Recall (Sensitivity), F1-score, Accuracy, and AUC—to provide a more reliable and comprehensive evaluation. These metrics help balance false positives and false negatives, which is critical in clinical applications where both misdiagnosis and missed detection carry risks.

\begin{table*}[!ht]
\centering
\caption{Performance Metrics (in \%) for Experiments E11 (left) and E12 (right)}
\label{tab-E11-E12}
\scriptsize
\begin{minipage}{0.42\textwidth}
\centering
\textbf{\hspace{2.8cm}Experiment E11}
\begin{tabular}{l|llllll}
\toprule
\textbf{Opt.} & \textbf{Model} & \textbf{Acc} & \textbf{PPV} & \textbf{Sen} & \textbf{F1} & \textbf{AUC} \\
\midrule
Hyperband & DT & 96.87 & 79.02 & 96.72 & 86.98 & 96.80 \\
& RF & 97.24 & 83.11 & 93.50 & 88.00 & 95.60 \\
& XGB & 97.55 & 86.15 & 92.19 & 89.07 & 95.20 \\
& DNN & 96.92 & 79.90 & 95.53 & 87.02 & 96.31 \\
& \textbf{RNN} & \textbf{97.18} & \textbf{81.74} & \textbf{95.19} & \textbf{96.31} & \textbf{87.95} \\
& SNN & 96.92 & 79.90 & 95.53 & 87.02 & 96.31 \\
\midrule
Optuna & DT & 97.55 & 85.03 & 93.83 & 89.22 & 95.92 \\
& RF & 97.51 & 85.28 & 93.04 & 88.99 & 95.55 \\
& \textbf{XGB} & \textbf{97.53} & \textbf{85.78} & \textbf{92.48} & \textbf{89.00} & \textbf{95.31} \\
& DNN & 97.16 & 84.35 & 90.55 & 87.34 & 94.26 \\
& RNN & 97.36 & 86.03 & 90.21 & 88.07 & 94.22 \\
& SNN & 96.84 & 79.24 & 95.87 & 86.77 & 96.41 \\
\midrule
GA & DT & 96.96 & 81.05 & 93.85 & 86.98 & 95.59 \\
& RF & 97.52 & 88.09 & 89.08 & 88.58 & 93.81 \\
& \textbf{XGB} & \textbf{97.68} & \textbf{86.88} & \textbf{92.48} & \textbf{89.59} & \textbf{95.39} \\
& DNN & 97.14 & 81.43 & 95.25 & 87.80 & 96.31 \\
& RNN & 96.91 & 79.47 & 96.32 & 87.09 & 96.65 \\
& SNN & 96.99 & 80.47 & 95.31 & 87.26 & 96.25 \\
\midrule
SMAC3 & DT & 97.57 & 86.06 & 92.32 & 89.08 & 95.26 \\
& RF & 97.43 & 84.56 & 93.12 & 88.63 & 95.53 \\
& \textbf{XGB} & \textbf{97.67} & \textbf{86.45} & \textbf{92.89} & \textbf{89.55} & \textbf{95.57} \\
& DNN & 97.05 & 82.68 & 91.81 & 87.01 & 94.75 \\
& RNN & 96.63 & 83.52 & 85.51 & 84.50 & 91.74 \\
& SNN & 77.40 & 32.74 & 99.83 & 49.31 & 87.23 \\
\midrule
\textbf{Without HPO} & RF & -- & 89.36 & 91.66 & 90.49 & -- \\
\bottomrule
\end{tabular}
\end{minipage}
\hfill
\begin{minipage}{0.42\textwidth}
\centering
\textbf{Experiment E12}
\begin{tabular}{l|llllll}
\toprule
 & \textbf{Model} & \textbf{Acc} & \textbf{PPV} & \textbf{Sen} & \textbf{F1} & \textbf{AUC} \\
\midrule
 & DT & 96.85 & 97.02 & 90.06 & 93.41 & 94.57 \\
& RF & 96.05 & 97.48 & 86.30 & 91.55 & 92.78 \\
& XGB & 96.98 & 97.32 & 90.30 & 93.68 & 94.74 \\
& DNN & 96.29 & 97.84 & 86.95 & 92.07 & 93.16 \\
& \textbf{RNN} & \textbf{96.98} & \textbf{97.71} & \textbf{89.89} & \textbf{94.60} & \textbf{93.64} \\
& SNN & 97.16 & 95.00 & 93.47 & 94.23 & 95.92 \\
\midrule
 & DT & 97.34 & 96.46 & 92.65 & 94.52 & 95.77 \\
& RF & 97.16 & 97.34 & 91.03 & 94.08 & 95.10 \\
& XGB & 96.98 & 97.32 & 90.29 & 93.68 & 94.74 \\
& \textbf{DNN} & \textbf{97.35} & \textbf{93.29} & \textbf{96.20} & \textbf{94.73} & \textbf{96.96} \\
& RNN & 97.15 & 97.38 & 90.95 & 94.05 & 95.07 \\
& SNN & 96.86 & 91.70 & 96.00 & 93.80 & 96.57 \\
\midrule
 & DT & 97.34 & 96.46 & 92.65 & 94.52 & 95.77 \\
& RF & 95.61 & 97.34 & 84.59 & 90.51 & 91.91 \\
& XGB & 97.40 & 96.93 & 92.45 & 94.64 & 95.74 \\
& DNN & 97.43 & 96.53 & 92.96 & 94.71 & 95.93 \\
& \textbf{RNN} & \textbf{97.54} & \textbf{96.57} & \textbf{93.39} & \textbf{94.95} & \textbf{96.15} \\
& SNN & 97.34 & 95.78 & 93.35 & 94.55 & 96.00 \\
\midrule
 & DT & 96.92 & 97.22 & 90.15 & 93.55 & 94.65 \\
& RF & 97.18 & 97.29 & 91.14 & 94.12 & 95.15 \\
& XGB & 96.95 & 97.29 & 90.20 & 93.61 & 94.69 \\
& DNN & 97.40 & 95.58 & 93.84 & 94.70 & 96.21 \\
& \textbf{RNN} & \textbf{97.49} & \textbf{95.49} & \textbf{94.32} & \textbf{94.90} & \textbf{96.42} \\
& SNN & 96.68 & 97.45 & 88.92 & 92.99 & 94.08 \\
\midrule
& RF & -- & 89.36 & 91.66 & 90.49 & -- \\
\bottomrule
\end{tabular}
\end{minipage}
\end{table*}


\begin{table*}[!ht]
\centering
\caption{Performance Metrics (in \%) for Experiments E21 (left) and E22 (right)}
\label{tab-E21-E22}
\scriptsize
\begin{minipage}{0.40\textwidth}
\centering
\textbf{\hspace{2.6cm}Experiment E21}
\begin{tabular}{l|llllll}
\toprule
\textbf{Opt.} & \textbf{Model} & \textbf{Acc} & \textbf{PPV} & \textbf{Sen} & \textbf{F1} & \textbf{AUC} \\
\midrule
Hyperband & DT  & 89.22 & 50.99 & 98.82 & 67.27 & 93.42 \\ 
   & RF  & 91.07 & 55.83 & 97.46 & 70.99 & 93.86 \\ 
   & XGB & 91.04 & 55.77 & 97.06 & 70.84 & 93.67 \\ 
   & DNN & 91.31 & 56.50 & 97.45 & 71.53 & 93.99 \\ 
   & {\bf RNN} & {\bf88.92} & {\bf50.30} & {\bf98.27} & {\bf93.01} & {\bf66.54} \\ 
   & SNN & 92.84 & 88.71 & 41.41 & 56.46 & 70.37 \\   
\midrule
Optuna & DT  & 89.22 & 50.99 & 98.82 & 67.27 & 93.42 \\ 
   & RF  & 90.82 & 55.09 & 97.72 & 70.46 & 93.83 \\ 
   & XGB & 91.28 & 56.47 & 96.97 & 71.38 & 93.77 \\ 
   & {\bf DNN} & {\bf 94.52} & {\bf 70.34} & {\bf 88.35} & {\bf 78.32} & {\bf 91.83} \\ 
   & RNN & 90.77 & 54.99 & 97.30 & 70.27 & 93.63 \\ 
   & SNN & 91.29 & 56.47 & 97.04 & 71.40 & 93.80 \\   
  \midrule
GA & DT  & 90.34 & 53.79 & 98.10 & 69.48 & 93.73 \\ 
   & RF  & 90.73 & 55.06 & 93.92 & 69.42 & 92.12 \\ 
   & {\bf XGB} & {\bf 94.10} & {\bf 68.52} & {\bf 87.61} & {\bf 76.90} & {\bf 91.27} \\ 
   & DNN & 92.50 & 60.41 & 95.95 & 74.14 & 94.01 \\ 
   & RNN & 91.40 & 56.87 & 96.25 & 71.50 & 93.52 \\ 
   & SNN & 92.24 & 59.46 & 96.55 & 73.59 & 94.12 \\   
   \midrule
SMAC3 & DT  & 91.28 & 56.47 & 96.80 & 71.33 & 93.69 \\ 
      & RF  & 90.79 & 55.01 & 97.84 & 70.42 & 93.87 \\ 
      & XGB & 91.20 & 56.24 & 96.98 & 71.19 & 93.73 \\ 
      & {\bf DNN} & {\bf 92.38} & {\bf 59.96} & {\bf 96.20} & {\bf 73.88} & {\bf 94.05} \\ 
      & RNN & 89.88 & 52.58 & 98.40 & 68.54 & 93.60 \\ 
      & SNN & 71.79 & 28.43 & 100.00 & 44.27 & 84.11 \\   
      \midrule
{\bf Without HPO} & SNN & -- &  64.36 & 93.78 & 76.34 & -- \\  
\bottomrule
\end{tabular}
\end{minipage}
\hfill
\begin{minipage}{0.40\textwidth}
\centering
\textbf{Experiment E22}
\begin{tabular}{l|llllll}
\toprule
 & \textbf{Model} & \textbf{Acc} & \textbf{PPV} & \textbf{Sen} & \textbf{F1} & \textbf{AUC} \\
\midrule
 & DT  & 98.18 & 96.33 & 96.59 & 96.46 & 97.66 \\ 
 & RF  & 98.18 & 96.67 & 96.22 & 96.44 & 97.54 \\ 
 & {\bf XGB} & {\bf 98.43} & {\bf 97.11} & {\bf 96.80} & {\bf 96.95} & {\bf 97.90} \\ 
 & DNN & 98.24 & 97.25 & 95.83 & 96.54 & 97.45 \\ 
 & RNN & 98.32 & 96.80 & 96.64 & 96.71 & 96.72\\ 
 & SNN & 80.49 & 99.84 & 23.97 & 38.66 & 61.98 \\   
 \midrule
 & DT  & 98.33 & 96.94 & 96.54 & 96.74 & 97.74 \\ 
 & RF  & 98.26 & 96.46 & 96.75 & 96.61 & 97.77 \\ 
 & {\bf XGB} & {\bf 98.34} & {\bf 96.96} & {\bf 96.57} & {\bf 96.77} & {\bf 97.76} \\ 
 & DNN & 98.30 & 96.93 & 96.44 & 96.68 & 97.69 \\ 
 & RNN & 98.30 & 96.36 & 97.03 & 96.69 & 97.88 \\ 
 & SNN & 98.31 & 96.97 & 96.44 & 96.70 & 97.70 \\   
 \midrule
 & DT  & 98.18 & 96.33 & 96.59 & 96.46 & 97.66 \\ 
 & RF  & 98.05 & 96.89 & 95.48 & 96.18 & 97.21 \\ 
 & XGB & 98.35 & 97.02 & 96.53 & 96.77 & 97.75 \\ 
 & DNN & 98.29 & 96.80 & 96.54 & 96.67 & 97.72 \\ 
 & {\bf RNN} & {\bf 98.31} & {\bf 96.96} & {\bf 96.43} & {\bf 96.69} & {\bf 97.69} \\ 
 & SNN & 98.10 & 97.75 & 94.78 & 96.24 & 97.01 \\   
 \midrule
 & DT  & 98.34 & 96.91 & 96.60 & 96.76 & 97.77 \\ 
 & RF  & 98.24 & 96.48 & 96.69 & 96.59 & 97.74 \\ 
 & {\bf XGB} & {\bf 98.37} & {\bf 96.99} & {\bf 96.66} & {\bf 96.83} & {\bf 97.81} \\ 
 & DNN & 98.27 & 96.69 & 96.55 & 96.62 & 97.71 \\ 
 & RNN & 98.20 & 97.01 & 96.03 & 96.52 & 97.50 \\ 
 & SNN & 96.31 & 96.49 & 88.89 & 92.53 & 93.88 \\   
 \midrule
 & CB & -- &  97.06 & 96.58  & 96.82 & -- \\  
\bottomrule
\end{tabular}
\end{minipage}
\end{table*}

The results, summarized in Tab.~\ref{tab-E11-E12}--\ref{tab-E21-E22}, highlight the effectiveness of HPO techniques in improving classification performance. For clarity, we bolded the best F1-score for each experiment–model pair, since F1 best reflects balanced performance under class imbalance.

In Experiment E11, GA performed well on accuracy and PPV with RF, while SMAC3 reached the highest sensitivity (99.83\%) with SNN but with very low PPV, showing the recall–precision trade-off. Hyperband achieved the best F1 and AUC, offering a more balanced outcome.

In E12, GA again excelled with the highest accuracy (97.54\%) and F1 (94.95\%) using RNN. Optuna maximized sensitivity (96.20\%) with DNN but slightly reduced PPV. SMAC3 and Hyperband maintained strong PPV, especially with DT and RF.

In E21, Optuna reached the highest accuracy (94.52\%) with DNN. Hyperband excelled in PPV and F1 with SNN, while SMAC3 achieved perfect sensitivity (100\%) but with low PPV, leading to many false positives.

In E22, Hyperband was the overall best, with the highest accuracy (98.43\%), F1 (96.95\%), and AUC (97.90\%) using XGB, though sometimes at the cost of sensitivity.

Across experiments, SMAC3 was the most stable, consistently balancing sensitivity, F1, and AUC, particularly with DL models. Hyperband and SMAC3 delivered the most competitive results overall—Hyperband often producing the top F1-scores, while SMAC3 provided robust, balanced performance across models.

\section{Discussion}
\label{discussion}


\subsection{Performance vs. Efficiency Trade-offs}


Among all experiments, E22 consistently demonstrated the best and most stable performance across all models. This makes it the most suitable setup for benchmarking. Within E22, the XGB classifier tuned with Hyperband achieved the highest performance overall, reaching an F1-score of 96.95\%. This highlights not only the strength of tree-based methods for AF detection but also the importance of effective hyperparameter optimization.

Hyperband proved to be the most efficient HPO method, combining high predictive performance with computational efficiency. By aggressively eliminating underperforming configurations early, Hyperband explores a wide hyperparameter space with reduced training cost. Compared to more resource-intensive methods like GA and Optuna, which may require more iterations or model evaluations, Hyperband achieves a strong balance between speed and model quality.

Table~\ref{tab-comp} presents a comparative overview of the best results obtained with each optimizer. Hyperband slightly outperformed GA, Optuna, and SMAC3 in all performance metrics, establishing it as the most effective HPO strategy for this task.

SMAC3 also demonstrated robust and stable performance across experiments, especially in terms of sensitivity and F1-score. While slightly slower than Hyperband, its model-based search strategy provides a reliable alternative when consistent performance is prioritized over speed.

\subsection{Effect of Hyperparameter Optimization Across Experiments}

\begin{figure}[!ht]
\centering
\begin{subfigure}[b]{0.49\textwidth}
    \includegraphics[width=\linewidth]{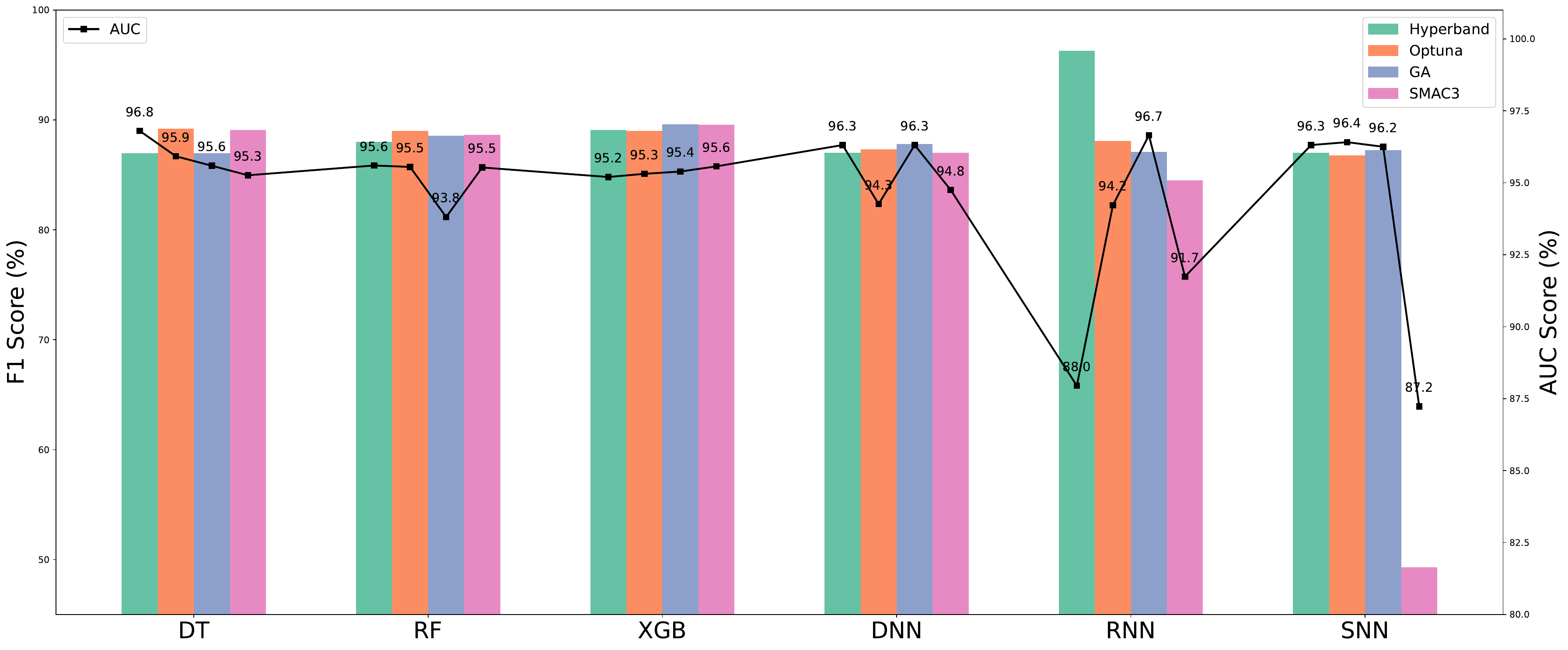}
    \caption{Results for E11}
    \label{fig:e11}
\end{subfigure}
\hfill
\begin{subfigure}[b]{0.49\textwidth}
    \includegraphics[width=\linewidth]{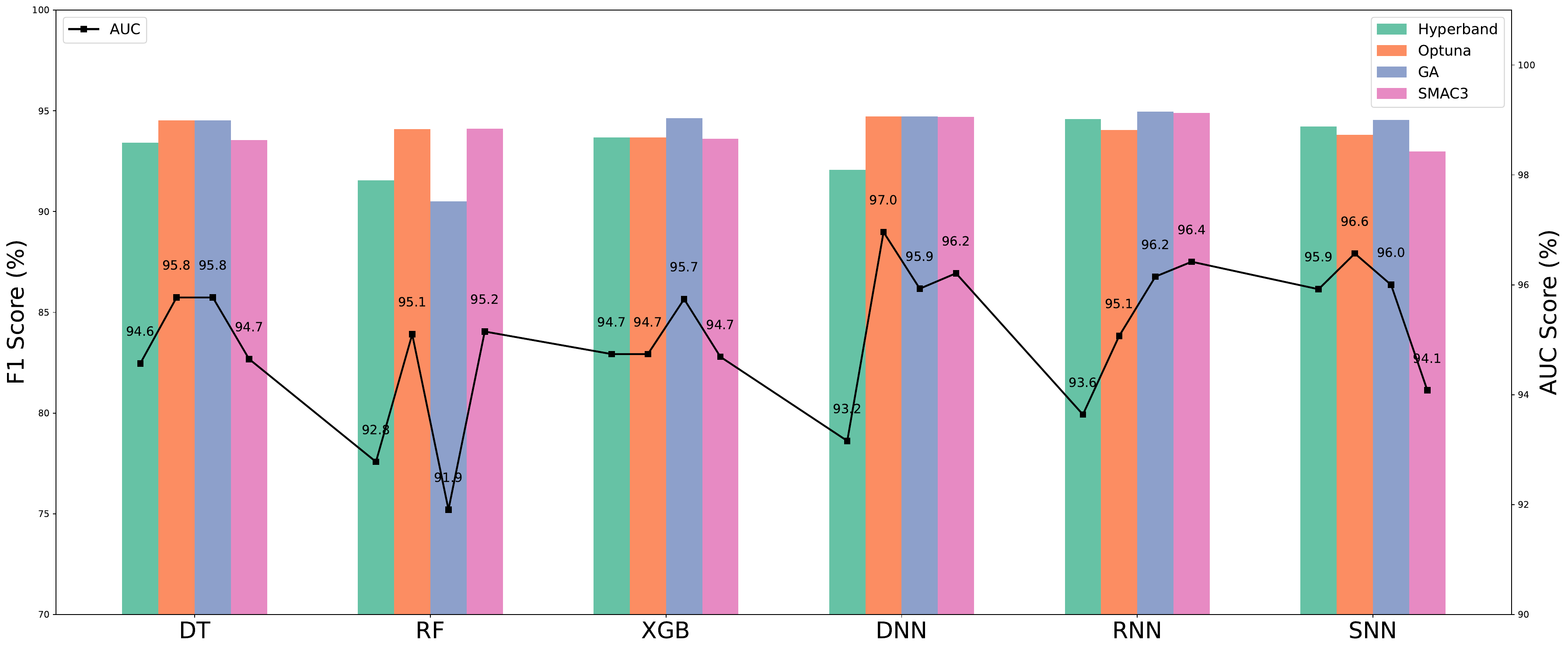}
    \caption{Results for E12}
    \label{fig:e12}
\end{subfigure}
\vskip\baselineskip
\begin{subfigure}[b]{0.49\textwidth}
    \includegraphics[width=\linewidth]{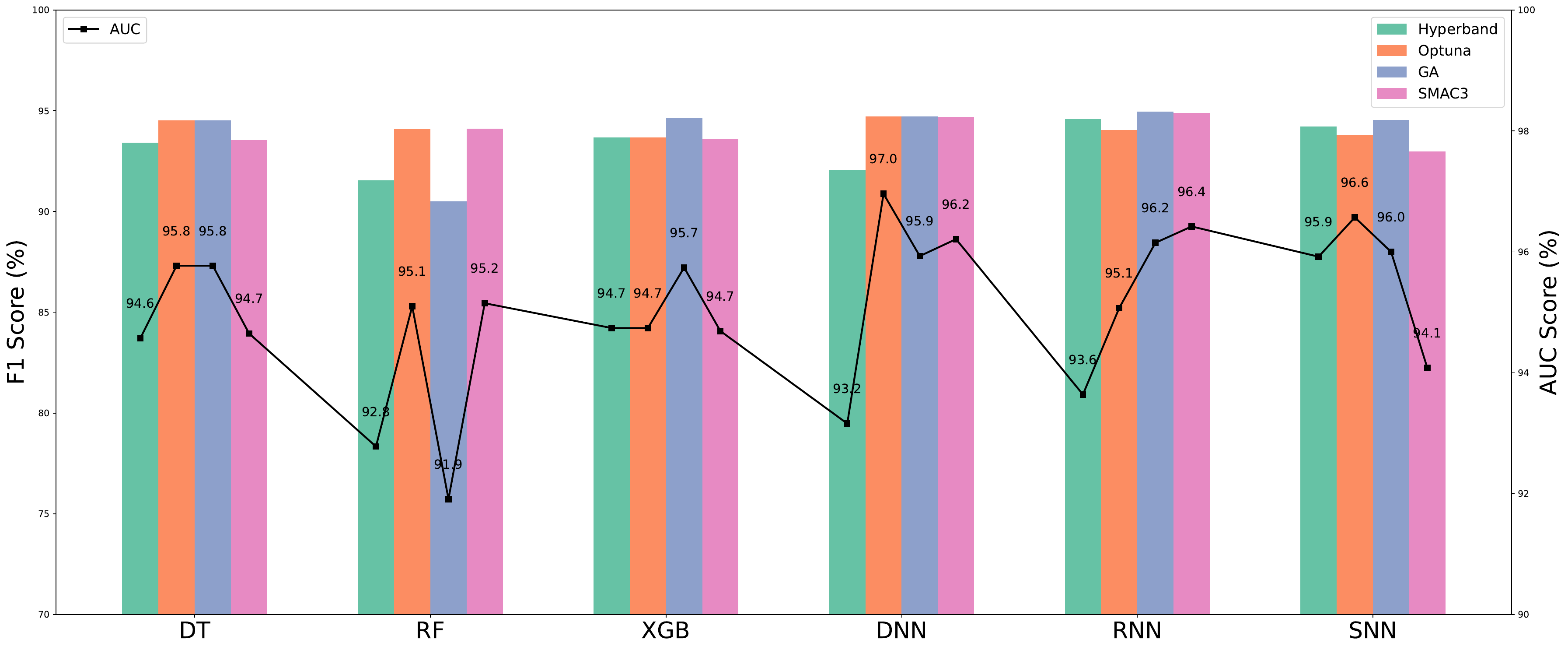}
    \caption{Results for E21}
    \label{fig:e21}
\end{subfigure}
\hfill
\begin{subfigure}[b]{0.49\textwidth}
    \includegraphics[width=\linewidth]{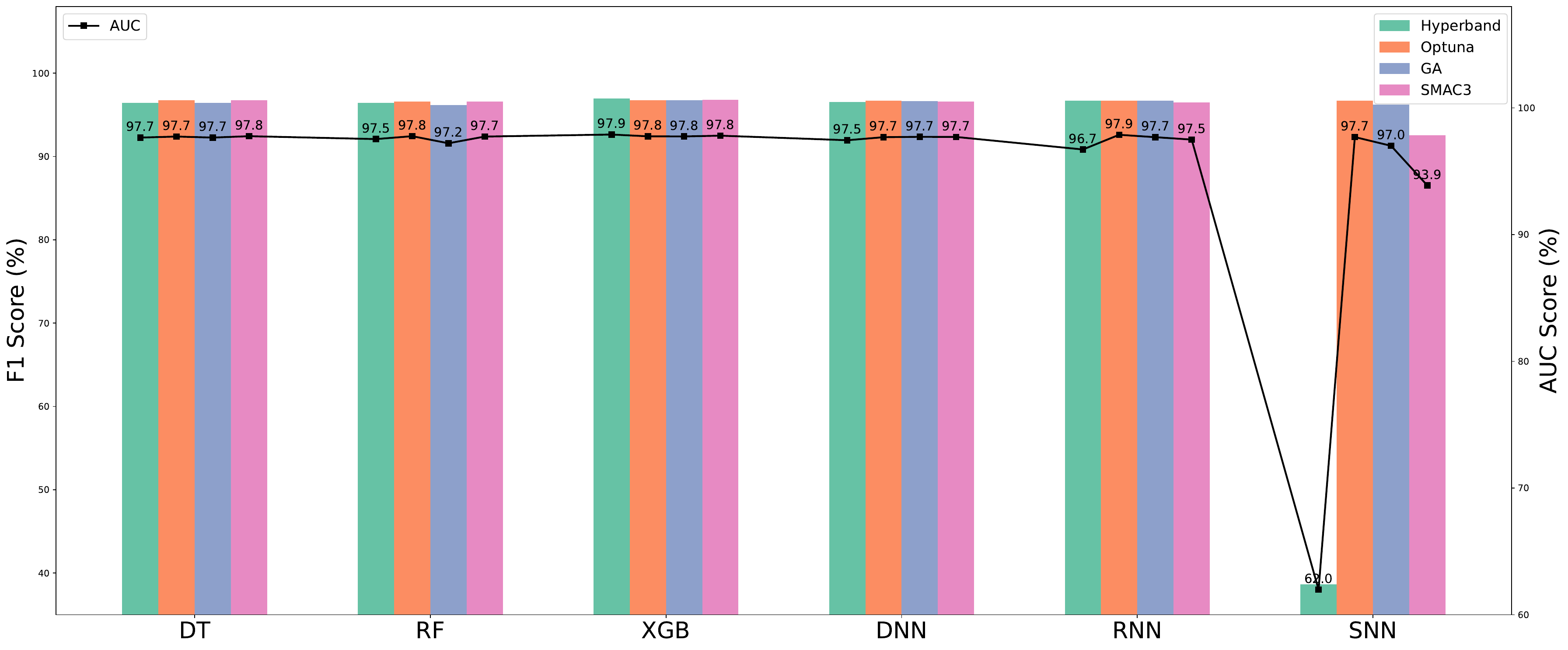}
    \caption{Results for E22}
    \label{fig:e22}
\end{subfigure}
\caption{Performance comparison for all experiments (E11, E12, E21, E22).}
\label{fig:all_experiments}
\end{figure}

Hyperparameter optimization led to substantial improvements across many models and experiments, highlighting its critical role in boosting performance. In particular as we can note from Fig.~\ref{fig:all_experiments}, Experiment E21 demonstrated the most significant gains. The RNN model tuned with Hyperband saw its F1-score jump from 72.61\% to 93.01\%, suggesting that recurrent models are highly sensitive to parameter tuning and benefit greatly from resource-aware optimization. Similarly, DNN with Optuna improved from 71.18\% to 78.32\%, and XGB with GA improved from 73.14\% to 76.90\%.

\begin{table}[!h]
\centering
\caption{The best results for E22}
\label{tab-comp}
\begin{tabular}{@{}lllllll@{}}
\toprule
 \textbf{Optimizer} & \textbf{Model} & \textbf{Acc} & \textbf{PPV} & \textbf{Sen} & \textbf{F1}  \\ 
\midrule
    GA & XGB & 98.35 & 97.02 & 96.53  & 96.77 \\    
    \midrule
    Hyperband & XGB & {\bf 98.43} & {\bf 97.11} & {\bf 96.80} & {\bf 96.95} \\    
    \midrule 
    Optuna & XGB & 98.34 & 96.96 & 96.57 & 96.77   \\   
    \midrule 
    SMAC3 & XGB & 98.37 & 96.99 & 96.66 & 96.83  \\ 
\hline
\end{tabular}
\end{table}

In Experiment E12, traditional models like DT and RF showed modest but consistent gains. DT improved from 93.85\% to 94.52\%, RF from 93.42\% to 94.08\% and XGB with GA reached 94.64\%. These results reinforce the effectiveness of GA for fine-tuning tree-based methods.

Even in high-performing settings like E22, minor improvements were observed. XGB with Hyperband improved slightly from 96.77\% to 96.95\% and XGB with SMAC3 reached 96.83\%. This demonstrates that even when models are near their performance ceiling, fine-tuning can achieve incremental gains.

However, not all optimization efforts resulted in improvement. For example, DT and RF in E21 underperformed after Hyperband tuning, and several models in E22 experienced slight drops. These setbacks may stem from overfitting to the validation set, poorly defined search spaces, or convergence to local optima. In scenarios where baseline models already perform well, aggressive tuning can introduce instability or noise, leading to marginal or negative changes.

Overall, the results suggest that while HPO is not universally beneficial, it can dramatically improve outcomes when model complexity, data distribution, and optimizer strategy are well-aligned.

\subsection{Comparison with State-of-the-Art Approaches}
\label{state_of_the_art_comparison}

Table~\ref{tab-comparison} compares our optimized models against several state-of-the-art methods for AF detection. 
Some studies applied Grid Search (GS), which systematically evaluates predefined parameter combinations. Other work used BOHB (Bayesian Optimization with Hyperband), which balances exploration and exploitation more efficiently. 

As opposed to these prior works, our study systematically applied and compared multiple HPO techniques to optimize both ML and DL models. As a result, our best-performing model—XGB tuned with Hyperband—achieved an F1-score of 96.95\% and an AUC of 97.90\%, outperforming most prior work in both predictive strength and robustness. Additionally, our inter-patient split setup ensures a more realistic generalization evaluation compared to the commonly used mixed-patient splits in other studies.

\begin{table}[!ht]
\centering
\caption{Comparison with state-of-the-art approaches}
\label{tab-comparison}
\begin{tabular}{@{}llllllllll@{}}
\toprule
\textbf{Ref.} & \textbf{Optimizer} & \textbf{Model} & \textbf{Method} &\textbf{Split}& \textbf{Acc} & \textbf{Sen} & \textbf{PPV} & \textbf{F1} &\textbf{AUC}  \\
\midrule
\cite{tudjarski2021new} & Optuna & XGB & mixed & 70/30 & - & - & - & 99.00 & - \\
    \midrule
    \cite{rawshani2024enhancing} & Optuna & AlexNet & mixed &  80/20 & - & - & - & - & 97.23  \\
    \midrule
    \cite{al2022detection} & GA & CNN & mixed & -& 99.95 & 100 & - & - & - \\
    \midrule
    \cite{ummadisetty2024atrial} & GS & SVM  & mixed & 60/40 & 96.50  & - & - & - & - \\
    \midrule
    \cite{han2020optimal} & BOHB & CNN/LSTM & mixed & 80/20 & 72.80 & - & - & 59.31 & - \\
    \midrule
    \cite{chen2024identifying} & Hyperband & DMNN & intra & dataset & - & - & - & - &  83.40 \\
    \midrule
    \textbf{our} & Hyperband & RNN & intra & 80/20 & 97.18  & 81.74  & 95.19 & 96.31 & 87.95 \\
    \midrule
    \textbf{our} & Hyperband & XGB & intra & 80/20 & 98.43  & 97.11  & 96.80 & 96.95 & 97.90 \\
    \midrule
    \textbf{our} & GA & RNN & inter & dataset & 97.54  & 96.57  & 93.39 & 94.95 & 96.15 \\
    \midrule
    \textbf{our} & Hyperband & RNN & inter & dataset & 88.92  & 50.30  & 98.27 & 93.01 & 66.54 \\
    
\hline
\end{tabular}
\end{table}

\subsection{Impact of Dataset Splitting on Generalization}

Dataset splitting strategy plays a critical role in evaluating model generalizability. Many prior studies use mixed or intra-patient splits, where data from the same individuals appear in both training and testing sets. While this often yields higher performance, it risks overfitting to patient-specific patterns.

In contrast, our study adopts a strict inter-patient split, ensuring no overlap between patients across training and testing. Although this typically leads to slightly lower metrics, it provides a more realistic and robust assessment of model performance.

The strong results achieved under this setup demonstrate the models’ ability to generalize across individuals—an essential requirement for clinical deployment and real world applicability.


\section{Conclusion}
\label{conclusion}
This study evaluated the impact of HPO techniques on AF detection using a range of ML and DL models. The results show that effective HPO significantly improves classification performance across both traditional and neural network-based models.

Among the tested combinations, XGBoost optimized with Hyperband under an inter-patient 80/20 split achieved the highest overall performance, with an AUC of 97.90\%, accuracy of 98.43\%, and sensitivity of 96.80\%. These results highlight strong predictive capability and generalization, especially in inter-patient settings that better reflect real-world clinical variability and reduce the risk of overfitting to patient-specific patterns.

The findings also emphasize the limitations of relying on single dataset training or mixed-patient splits, which may produce overly optimistic results. In contrast, our use of inter-dataset and inter-patient evaluations provides a more robust assessment of model stability and generalizability.

Notably, we introduced SMAC3—a rarely used HPO method in AF detection—which demonstrated consistent performance and shows promise for broader application in medical time-series optimization tasks.

For future work, we plan to expand the evaluation to more diverse ECG datasets and explore domain adaptation and transfer learning techniques. Additionally, developing ensemble approaches and resource-efficient models will be critical to supporting real-time, clinically deployable AF detection systems.

\balance
\bibliographystyle{unsrt}  
\bibliography{references}  

\end{document}